\documentclass[conference]{IEEEtran}
\IEEEoverridecommandlockouts
\usepackage{cite}
\usepackage{amsmath,amssymb,amsfonts}
\usepackage{algorithm}
\usepackage[noend]{algpseudocode}
\usepackage{graphicx}
\usepackage{textcomp}
\usepackage{xcolor}
\usepackage[bookmarks=false]{hyperref}       
\usepackage{url}            
\usepackage{float}

\ifCLASSOPTIONcompsoc
    \usepackage[caption=false,font=normalsize,labelfont=sf,textfont=sf]{subfig}
\else
    \usepackage[caption=false,font=footnotesize]{subfig}
\fi

\def\BibTeX{{\rm B\kern-.05em{\sc i\kern-.025em b}\kern-.08em
    T\kern-.1667em\lower.7ex\hbox{E}\kern-.125emX}}

\begin{document}

\title{A general representation of dynamical systems for reservoir computing}

\author{\IEEEauthorblockN{Sidney Pontes-Filho\IEEEauthorrefmark{1}\textsuperscript{,}\IEEEauthorrefmark{2}\textsuperscript{,}\IEEEauthorrefmark{4}, Anis Yazidi\IEEEauthorrefmark{1}, Jianhua Zhang\IEEEauthorrefmark{1}, Hugo Hammer\IEEEauthorrefmark{1},\\Gustavo B. M. Mello\IEEEauthorrefmark{1}, Ioanna Sandvig\IEEEauthorrefmark{3}, Gunnar Tufte\IEEEauthorrefmark{2} and Stefano Nichele\IEEEauthorrefmark{1}}
\IEEEauthorblockA{\IEEEauthorrefmark{1}\textit{Department of Computer Science, Oslo Metropolitan University, Oslo, Norway}\\
\IEEEauthorrefmark{2}\textit{Department of Computer Science, Norwegian University of Science and Technology, Trondheim, Norway}\\
\IEEEauthorrefmark{3}\textit{Department of Neuromedicine and Movement Science, Norwegian University of Science and Technology, Trondheim, Norway}\\
Email: \IEEEauthorrefmark{4}sidneyp@oslomet.no}}

\maketitle

\begin{abstract}
Dynamical systems are capable of performing computation in a reservoir computing paradigm. This paper presents a general representation of these systems as an artificial neural network (ANN). Initially, we implement the simplest dynamical system, a cellular automaton. The mathematical fundamentals behind an ANN are maintained, but the weights of the connections and the activation function are adjusted to work as an update rule in the context of cellular automata. The advantages of such implementation are its usage on specialized and optimized deep learning libraries, the capabilities to generalize it to other types of networks and the possibility to evolve cellular automata and other dynamical systems in terms of connectivity, update and learning rules. Our implementation of cellular automata constitutes an initial step towards a general framework for dynamical systems. It aims to evolve such systems to optimize their usage in reservoir computing and to model physical computing substrates.
\end{abstract}


\section{Introduction}
\label{sec:intro}

A cellular automaton (CA) is the simplest computing system where the emergence of complex dynamics from local interactions might take place. It consists of a grid of cells with a finite number of states that change according to simple rules depending on the neighborhood and own state in discrete time-steps. Some notable examples are the elementary CA\cite{wolfram2002new}, which is unidimensional with three neighbors and eight update rules, and Conway's Game of Life\cite{Rendell2002}, which is two-dimensional with nine neighbors and three update rules.

Table~\ref{table:dynamical_systems} presents some computing systems that are capable of giving rise to the emergence of complex dynamics. Those systems can be exploited by reservoir computing which is a paradigm that resorts to dynamical systems to simplify complex data. Hence, simpler and faster machine learning methods can be applied with such simplified data. Reservoir computing is more energy efficient than deep learning methods and it can even yield competitive results, especially for temporal data \cite{schrauwen2007overview}. In short, reservoir computing exploits a dynamical system that possesses the echo state property and fading memory, where the internals of the reservoir are untrained and the only training happens at the linear readout stage \cite{Konkoli2018}. Reservoir computers are most useful when the substrate's dynamics are at the ``edge of chaos", meaning a range of dynamical behaviors that is between order and disorder \cite{LANGTON199012}. Cellular automata with such dynamical behavior are capable of being exploited as reservoirs \cite{Nichele2017reservoir,Nichele2017deeplearning}. Other systems can also exhibit the same dynamics. The coupled map lattice \cite{kaneko1992overview} is very similar to CA, the only exception is that the coupled map lattice has continuous states which are updated by a recurrence equation involving the neighborhood. Random Boolean network \cite{gershenson2004introduction} is a generalization of CA where random connectivity exists. Echo state network \cite{Jaeger78} is an artificial neural network (ANN) with random topology while liquid state machine \cite{MAASS2004593} is similar to echo state network with the difference that it is a spiking neural network that communicates through discrete-events (spikes) over continuous time. 
One important aspect of the computation performed in a dynamical system is the trajectory of system's states traversed during the computation \cite{Nichele2010trajectories}. Such trajectory may be guided by system parameters \cite{Nichele2012Genome}. Computation in dynamical systems may be carried out in physical substrates \cite{TANAKA2019100}, such as networks of biological neurons \cite{aaser2017cyborg} or in other nanoscale materials \cite{Broersma2017}. Finding the correct abstraction for the computation in a dynamical system, e.g. CA, is an open problem \cite{nichele2017universality}. All the systems described in Table~\ref{table:dynamical_systems} are sparsely connected and can be represented by an adjacency matrix, such as a graph. A fully connected feedforward ANN represents its connectivity from a layer to another with an adjacency matrix that contains the weights of each connection. Our CA implementation is similar to this, but the connectivity is from the "layer" of cells to itself.

\begin{table}[t]
\renewcommand{\arraystretch}{1.1}
\caption{Examples of dynamical systems.}
\label{table:dynamical_systems}
\begin{tabular}{|l|l|l|l|}
\hline
\textbf{Dynamical system} & \textbf{State} & \textbf{Time} & \textbf{Connectivity} \\ \hline
Cellular automata & Discrete & Discrete & Regular \\ \hline
Coupled map lattice & Continuous & Discrete & Regular \\ \hline
Random Boolean network & Discrete & Discrete & Random \\ \hline
Echo state network & Continuous & Discrete & Random \\ \hline
Liquid state machine & Discrete & Continuous & Random \\ \hline
\end{tabular}%
\end{table}

The goal of representing CA with an adjacency matrix is to implement a framework which facilitates the development of all types of CAs, from unidimensional to multidimensional, with all kinds of lattices and without any boundary checks during execution; and also the inclusion of the major dynamical systems, independent of the type of the state, time and connectivity. Such initial implementation is the first part of a Python framework under development, based on TensorFlow deep neural network library \cite{abadi2016tensorflow}. Therefore, it benefits from powerful and parallel computing systems with multi-CPU and multi-GPU. This framework, called EvoDynamic\footnote{EvoDynamic v0.1 available at \url{https://github.com/SocratesNFR/EvoDynamic}.}, aims at evolving the connectivity, update and learning rules of sparsely connected networks to improve their usage for reservoir computing guided by the echo state property, fading memory, state trajectory and other quality measurements, and to model the dynamics and behavior of physical reservoirs, such as \textit{in-vitro} biological neural networks interfaced with microelectrode arrays and nanomagnetic ensembles. Those two substrates have real applicability as reservoirs. For example, the former substrate is applied to control a robot, in fact making it into a cyborg, a closed-loop biological-artificial neuro-system \cite{aaser2017cyborg}, and the latter possesses computation capability as shown by a square lattice of nanomagnets \cite{Jensen2018artificial}. Those substrates are the main interest of the SOCRATES project \cite{socratesweb} which aims to explore a dynamic, robust and energy efficient hardware for data analysis.

There are some implementations of CA similar to the one of EvoDynamic framework. They normally implement Conway's Game of Life by applying 2D convolution with a kernel that is used to count the neighbors, then the resulting matrix consists of the number of neighboring cells and is used to update the CA. One such implementation, also based on TensorFlow, is available open-source in \cite{conv2d}.

This paper is organized as follows. Section~\ref{sec:method} describes our method according to which we use adjacency matrix to compute CA. Section~\ref{sec:results} presents the results obtained from the method. Section~\ref{sec:future} discusses the future plan of EvoDynamic framework and Section~\ref{sec:conclusion} concludes this paper.


\section{Method}
\label{sec:method}

In our proposed method, the equation to calculate the next states of the cells in a cellular automaton is 

\begin{equation}
\label{eq0}
\mathbf{ca}_{t+1} = f(\mathbf{A}\cdot \mathbf{ca}_{t}).
\end{equation}

It is similar to the equation of the forward pass of an artificial neural network, but without the bias. The layer is connected to itself, and the activation function $f$ defines the update rules of the CA. The next states of the CA $\mathbf{ca}_{t+1}$ is calculated from the result of the activation function $f$ which receives as argument the dot product between the adjacency matrix $\mathbf{A}$ and the current states of the CA $\mathbf{ca}_t$. $\mathbf{ca}$ is always a column vector of size $len(\mathbf{ca})\times 1$, that does not depend on how many dimensions the CA has, and $\mathbf{A}$ is a matrix of size $len(\mathbf{ca})\times len(\mathbf{ca})$. Hence the result of $\mathbf{A}\cdot \mathbf{ca}$ is also a column vector of size $len(\mathbf{ca})\times 1$ as $\mathbf{ca}$.

The implementation of cellular automata as an artificial neural network requires the procedural generation of the adjacency matrix of the grid. In this way, any lattice type or multidimensional CAs can be implemented using the same approach. The adjacency matrix of a sparsely connected network contains many zeros because of the small number of connections. Since we implement it on TensorFlow, the data type of the adjacency matrix is preferably a \verb#SparseTensor#. A dot product with this data type can be up to 9 times faster depending on the configuration of the tensors \cite{tfsparse}. The update rule of the CA alters the weights of the connections in the adjacency matrix. In a CA whose cells have two states meaning ``dead" (zero) or ``alive" (one), the weights in the adjacency matrix are one for connection and zero for no connection, such as an ordinary adjacency matrix. Such matrix facilitates the description of the update rule for counting the number of ``alive" neighbors because the result of the dot product between the adjacency matrix and the cell state vector is the vector that contains the number of ``alive" neighbors for each cell. If the pattern of the neighborhood matters in the update rule, each cell has its neighbors encoded as a $n$-ary string where $n$ means the number of states that a cell can have. In this case the weights of the connections with the neighbors are $n$-base identifiers and are calculated by

\begin{equation}
\label{eq1}
neighbor_{i}=n^{i},\forall {i}\in \{0..len(\mathbf{neighbors})-1\}.
\end{equation}

Where $\mathbf{neighbors}$ is a vector of the cell's neighbors. In the adjacency matrix, each neighbor receives a weight according to \eqref{eq1}. The result of the dot product with such adjacency matrix is a vector that consists of unique integers per neighborhood pattern. Thus, the activation function is a lookup table from integer (i.e., pattern) to next state.

Algorithm~\ref{algo:gen1d} generates the adjacency matrix for one-dimensional CA, such as the elementary CA. Where $widthCA$ is the width or number of cells of a unidimensional CA and $\mathbf{neighborhood}$ is a vector which describes the region around the center cell. The connection weights depend on the type of update rule as previously explained. For example, in case of an elementary CA $\mathbf{neighborhood}=[4\ 2\ 1]$. $indexNeighborCenter$ is the index of the center cell in the $\mathbf{neighborhood}$ whose starting index is zero. $isWrappedGrid$ is a Boolean value that works as a flag for adding wrapped grid or not. A wrapped grid for one-dimensional CA means that the initial and final cells are neighbors. With all these parameters, Algorithm~\ref{algo:gen1d} creates an adjacency matrix by looping over the indices of the cells (from zero to $numberOfCells-1$) with an inner loop for the indices of the neighbors. If the selected $currentNeighbor$ is a non-zero value and its indices do not affect the boundary condition, then the value of $currentNeighbor$ is assigned to the adjacency matrix $\mathbf{A}$ in the indices that correspond to the connection between the current cell in the outer loop and the actual index of $currentNeighbor$. Finally, this procedure returns the adjacency matrix $\mathbf{A}$.

\begin{algorithm*}[ht]
\caption{Generation of adjacency matrix for 1D cellular automaton}
\label{algo:gen1d}
\begin{algorithmic}[1]
    \Procedure{generateCA1D}{$widthCA, \mathbf{neighborhood}, indexNeighborCenter, isWrappedGrid$}
    \State $numberOfCells\gets widthCA$
    \State $\mathbf{A}\gets \mathbf{0}^{numberOfCells\times numberOfCells}$
    \Comment{Adjacency matrix initialization}
    \For{$i\gets \{0..numberOfCells-1\}$}
    \For{$j\gets \{-indexNeighborCenter..len(neighborhood)-indexNeighborCenter-1\}$}
    \State $currentNeighbor\gets neighborhood_{j+indexNeighborCenter}$
    \If{$currentNeighbor\neq 0 \land (isWrappedGrid\lor (\lnot isWrappedGrid \land (0\leq(i+j)<widthCA))$}
    \State $\mathbf{A}_{i,((i+j)\bmod widthCA)}\gets currentNeighbor$
    \EndIf
    \EndFor
    \EndFor
    \State \Return $\mathbf{A}$
    \EndProcedure
\end{algorithmic}
\end{algorithm*}

To procedurally generate an adjacency matrix for 2D CA instead of 1D CA, the algorithm needs to have small adjustments. Algorithm~\ref{algo:gen2d} shows that for two-dimensional CA, such as Conway's Game of Life. In this case, the height of the CA is an argument passed as $heightCA$. $\mathbf{Neighborhood}$ is a 2D matrix and $\mathbf{indexNeighborCenter}$ is a vector of two components meaning the indices of the center of $\mathbf{Neighborhood}$. This procedure is similar to the one in Algorithm~\ref{algo:gen1d}, but it contains one more loop for the additional dimension.

\begin{algorithm*}
\caption{Generation of adjacency matrix of 2D cellular automaton}
\label{algo:gen2d}
\begin{algorithmic}[1]
    \Procedure{generateCA2D}{$widthCA, heightCA, \mathbf{Neighborhood}, \mathbf{indexNeighborCenter}, isWrappedGrid$}
    \State $numberOfCells\gets widthCA*heightCA$
    \State $\mathbf{A}\gets \mathbf{0}^{numberOfCells\times numberOfCells}$
    \Comment{Adjacency matrix initialization}
    \State $widthNB, heightNB\gets shape(\mathbf{Neighborhood})$
    \For{$i\gets \{0..numberOfCells-1\}$}
    \For{$j\gets \{-\mathbf{indexNeighborCenter}_0..widthNB-\mathbf{indexNeighborCenter}_0-1\}$}
    \For{$k\gets \{-\mathbf{indexNeighborCenter}_1..heightNB-\mathbf{indexNeighborCenter}_1-1\}$}
    \State $currentNeighbor\gets Neighborhood_{j+indexNeighborCenter}$
    \If{$currentNeighbor\neq 0 \land (isWrappedGrid\lor (\lnot isWrappedGrid \land (0\leq((i\bmod heightCA)+j)<widthCA) \land (0\leq(\lfloor i/widthCA\rfloor+k)<heightCA))$}
    \State $\mathbf{A}_{i,(((i+k)\bmod widthCA)+((\lfloor i/widthCA\rfloor+j)\bmod heightCA)*widthCA)}\gets currentNeighbor$
    \EndIf
    \EndFor
    \EndFor
    \EndFor
    \State \Return $\mathbf{A}$
    \EndProcedure
\end{algorithmic}
\end{algorithm*}

The activation function for CA is different from the ones used for ANN. For CA, it contains the update rules that verify the vector returned by the dot product between the adjacency matrix and the vector of states. Normally, the update rules of the CA are implemented as a lookup table from neighborhood to next state. In our implementation, the lookup table maps the resulting vector of the dot product to the next state of the central cell.

\section{Results}
\label{sec:results}

This section presents the results of the proposed method and it also stands for the preliminary results of the EvoDynamic framework.

\begin{figure}
\centering
\subfloat[]{\label{fig:ca1dA}\includegraphics[width=0.48\textwidth]{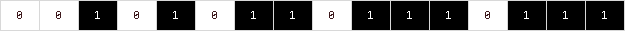}}\\
\subfloat[]{\label{fig:ca1dB}\includegraphics[width=0.48\textwidth]{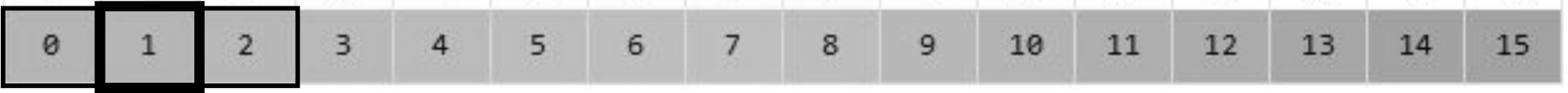}}\\
\subfloat[]{\label{fig:ca1dC}\includegraphics[width=0.48\textwidth]{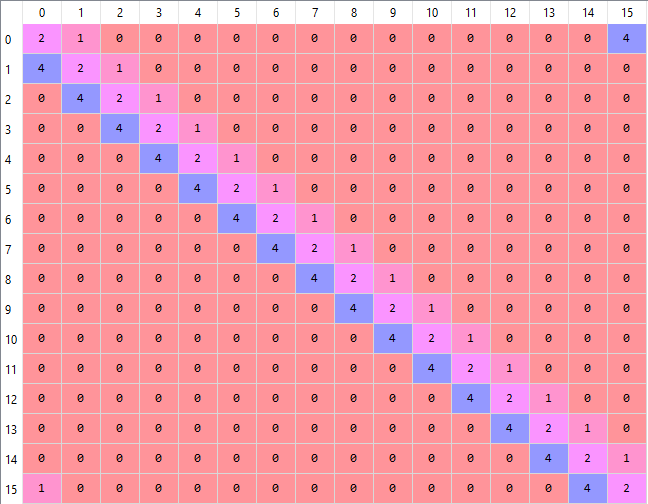}}
\caption{Elementary cellular automaton with 16 cells and wrapped grid. (a) Example of the grid of cells with states. (b) Indices of the cells and standard pattern neighborhood of elementary CA where thick border means the central cell and thin border means the neighbors. (c) Generated adjacency matrix for this elementary CA.}
\label{fig:ca1d}
\end{figure}

Fig.~\ref{fig:ca1d} illustrates a wrapped elementary CA described in the procedure of Algorithm~\ref{algo:gen1d} and its generated adjacency matrix. Fig.~\ref{fig:ca1dA} shows the appearance of the desired elementary CA with 16 cells (i.e., $widthCA=16$). Fig.~\ref{fig:ca1dB} describes its pattern 3-neighborhood and the indices of the cells. Fig~\ref{fig:ca1dC} shows the result of the Algorithm~\ref{algo:gen1d} with the neighborhood calculated by \eqref{eq1} for pattern matching in the activation function. In Fig.~\ref{fig:ca1dC}, we can verify that the left neighbor has weight equals to 4 (or $2^2$ for the most significant bit), central cell weight is 2 (or $2^1$) and right neighbor weight is 1 (or $2^0$ for the least significant bit) as defined by \eqref{eq1}. Since the CA is wrapped, we can notice in row index 0 of the adjacency matrix in Fig.~\ref{fig:ca1dC} that the left neighbor of cell 0 is the cell 15, and in row index 15 that the right neighbor of cell 15 is the cell 0.

\begin{figure}
\centering
\subfloat[]{\label{fig:ca2dA}\includegraphics[width=0.12\textwidth]{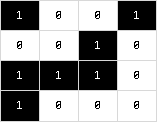}}
\hspace{0.1\textwidth}
\subfloat[]{\label{fig:ca2dB}\includegraphics[width=0.12\textwidth]{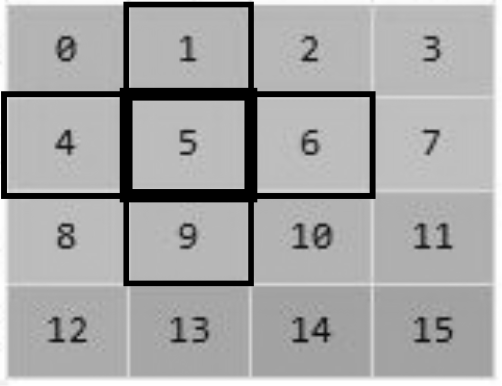}}\\
\subfloat[]{\label{fig:ca2dC}\includegraphics[width=0.48\textwidth]{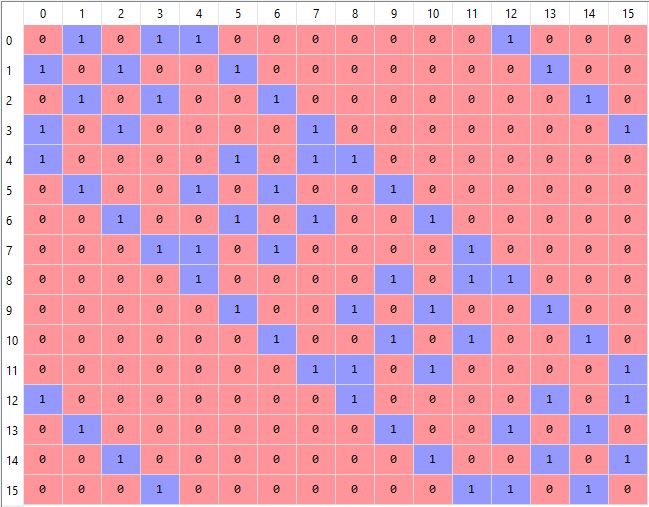}}
\caption{2D cellular automaton with 16 cells ($4\times 4$) and wrapped grid. (a) Example of the grid of cells with states. (b) Indices of the cells and von Neumann counting neighborhood of 2D CA where thick border means the current cell and thin border means the neighbors. (c) Generated adjacency matrix for this 2D CA.}
\label{fig:ca2d}
\end{figure}

Fig.~\ref{fig:ca2d} describes a wrapped 2D CA for Algorithm~\ref{algo:gen2d} and shows the resulting adjacency matrix. Fig.~\ref{fig:ca2dA} illustrates the desired two-dimensional CA with 16 cells (i.e., $widthCA=4$ and $heightCA=4$). Fig.~\ref{fig:ca2dB} presents the von Neumann neighborhood \cite{toffoli1987cellular} which is used for counting the number of "alive" neighbors (the connection weights are only zero and one, and $\mathbf{Neighborhood}$ argument of Algorithm~\ref{algo:gen2d} defines it). It also shows the index distribution of the CA whose order is preserved after flatting it to a column vector. Fig~\ref{fig:ca2dC} contains the generated adjacency matrix of Algorithm~\ref{algo:gen2d} for the described 2D CA. Fig.~\ref{fig:ca2dB} shows an example of a central cell with its neighbors, the index of this central cell is 5 and the row index 5 in the adjacency matrix of Fig.~\ref{fig:ca2dC} presents the same neighbor indices, i.e., 1, 4, 6 and 9. Since this is a symmetric matrix, the columns have the same connectivity of the rows. Therefore, this adjacency matrix represents an undirected graph. The wrapping effect is also observable. For example, the neighbors of the cell index 0 are 1, 3, 4 and 12. So the neighbors 3 and 12 are the ones that the wrapped grid allowed to exist for cell index 0.

\section{EvoDynamic future}
\label{sec:future}

\begin{figure*}
\centering
\subfloat[Step 1]{\label{fig:pca01}\includegraphics[width=0.22\textwidth]{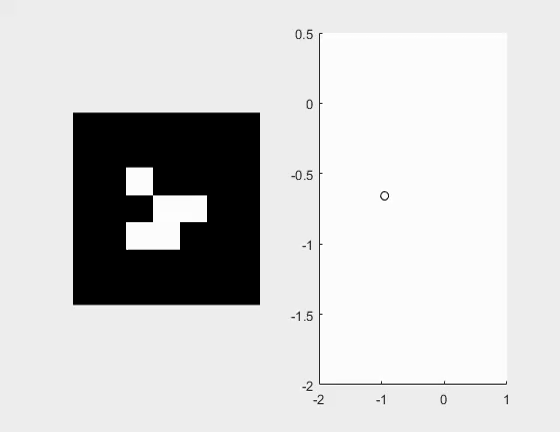}}\hfill
\subfloat[Step 2]{\label{fig:pca02}\includegraphics[width=0.22\textwidth]{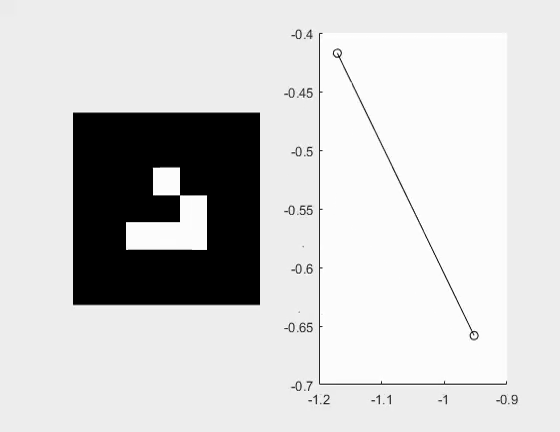}}\hfill
\subfloat[Step 3]{\label{fig:pca03}\includegraphics[width=0.22\textwidth]{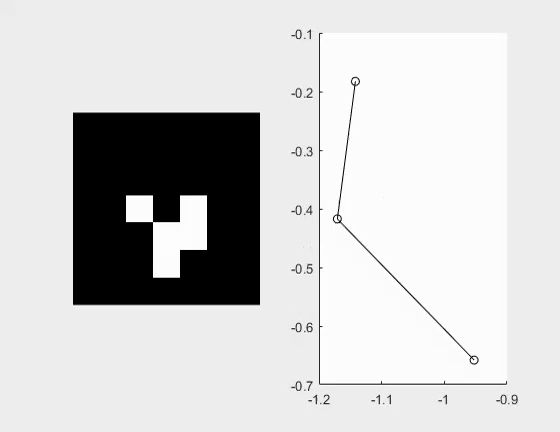}}\hfill
\subfloat[Step 4]{\label{fig:pca04}\includegraphics[width=0.22\textwidth]{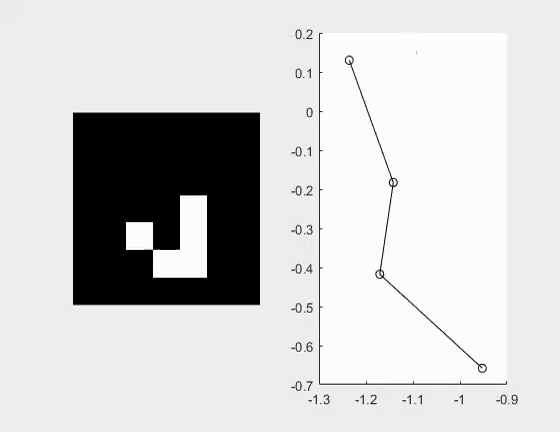}}\\
\subfloat[Step 11]{\label{fig:pca11}\includegraphics[width=0.22\textwidth]{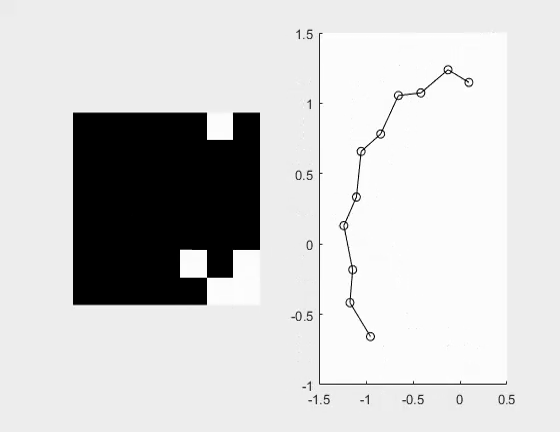}}\hfill
\subfloat[Step 12]{\label{fig:pca12}\includegraphics[width=0.22\textwidth]{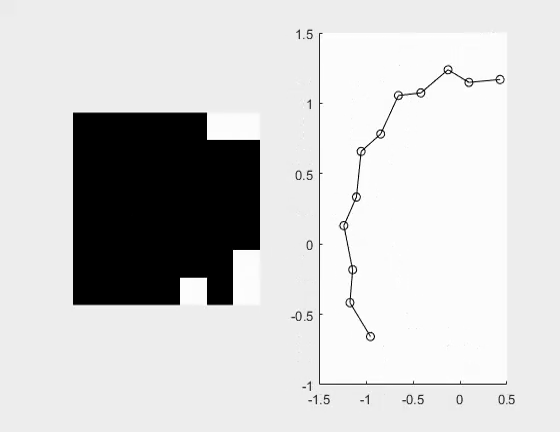}}\hfill
\subfloat[Step 13]{\label{fig:pca13}\includegraphics[width=0.22\textwidth]{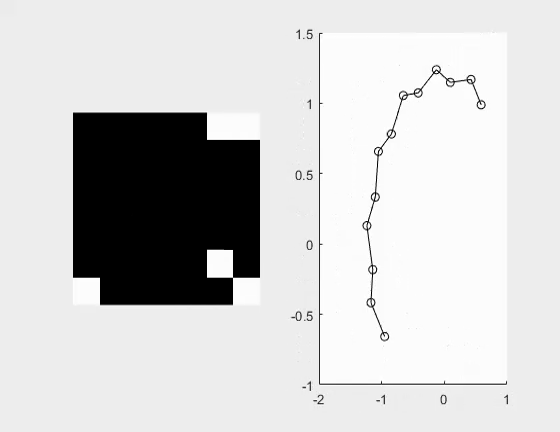}}\hfill
\subfloat[Step 14]{\label{fig:pca14}\includegraphics[width=0.22\textwidth]{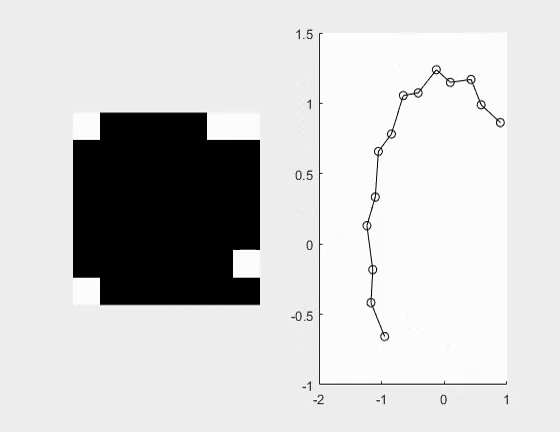}}\\
\subfloat[Step 26]{\label{fig:pca26}\includegraphics[width=0.22\textwidth]{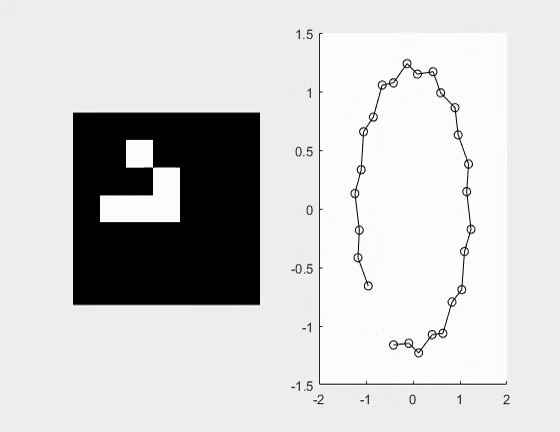}}\hfill
\subfloat[Step 27]{\label{fig:pca27}\includegraphics[width=0.22\textwidth]{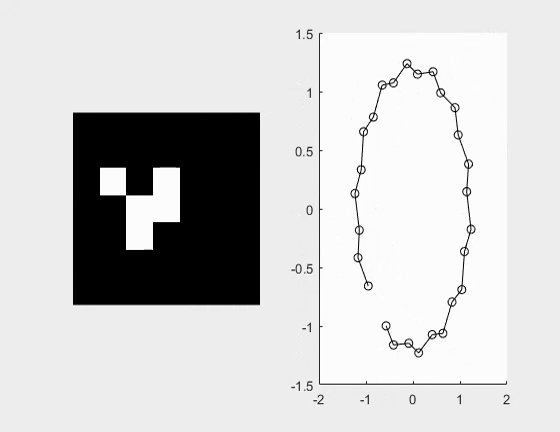}}\hfill
\subfloat[Step 28]{\label{fig:pca28}\includegraphics[width=0.22\textwidth]{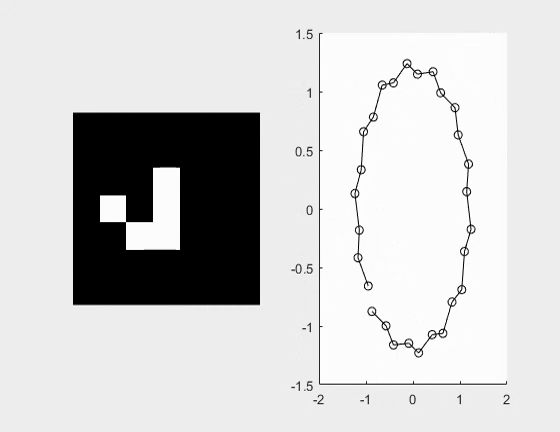}}\hfill
\subfloat[Step 29]{\label{fig:pca29}\includegraphics[width=0.22\textwidth]{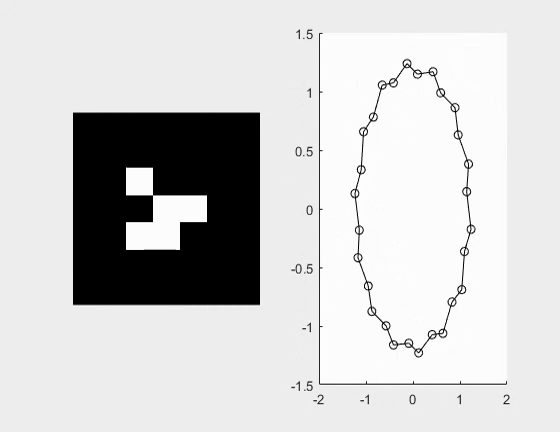}}
\caption{States of Conway's Game of Life in a 7x7 wrapped lattice alongside their PCA-transformed state transition diagrams of the two first principal components. (a) Initial state is a glider. (a)-(d) Four first steps in this CA. (e)-(h) Four intermediate steps in this CA while reaching the wrapped border. (i)-(l) Four last steps in this CA before repeating the initial state and closing a cycle.}
\label{fig:pca}
\end{figure*}

The method of implementing a CA as an artificial neural network will be beneficial for the future of EvoDynamic framework. Since the implementation of all sparsely connected networks in Table~\ref{table:dynamical_systems} are already planned in future releases of the Python framework, EvoDynamic must have a general representation to all of them. Therefore we are treating CA as an ANN. Moreover, EvoDynamic framework will evolve the connectivity, update and learning rules of the dynamical systems for reservoir computing improvement and physical substrate modeling. This common representation facilitates the evolution of such systems and models which will be guided by several methods that measure the quality of a reservoir or the similarity to a dataset. One example of these methods is the state trajectory. For visualization, we use principal component analysis (PCA) to reduce the dimensionality of the states and present them as a state transition diagram as shown in Fig.~\ref{fig:pca}.

\section{Conclusion}
\label{sec:conclusion}

In this paper, we present an alternative method to implement a cellular automaton. This allows any CA to be computed as an artificial neural network. Therefore, this will help to extend the CA implementation to more complex dynamical systems, such as echo state networks and liquid state machines. Furthermore, the EvoDynamic framework is built on a deep learning library, TensorFlow, which permits the acceleration of the execution when applied on parallel computational platforms with fast CPUs and GPUs. The future work for this CA implementation is to develop algorithms to procedurally generate adjacency matrices for 3D and multidimensional cellular automata with different types of cells, such as the cells with hexagonal shape.

\section*{Acknowledgments}
This work was supported by Norwegian Research Council SOCRATES project (grant number 270961).

\bibliographystyle{IEEEtran}
\bibliography{IEEEabrv,bibexample}

\end{document}